\documentclass[runningheads]{llncs}
\usepackage[T1]{fontenc}
\usepackage{graphicx}
\usepackage{amssymb}
\usepackage{amsmath}
\usepackage{url}
\usepackage{multirow}
\usepackage{float}
\usepackage{subfig}
\usepackage{array}
\usepackage{tabularx}
\usepackage{graphicx}
\usepackage{color,array}
\usepackage{comment}
\usepackage{adjustbox} 
\usepackage{makecell} 
\usepackage{arydshln} 
\usepackage[numbers]{natbib}
\usepackage{hyperref}
\usepackage[utf8]{inputenc}

\DeclareUnicodeCharacter{015F}{\c{s}}
\DeclareUnicodeCharacter{015F}{\c{s}}  

\DeclareUnicodeCharacter{011F}{\u{g}}
\DeclareUnicodeCharacter{0130}{\.{I}} 

\newcommand{\orcid}[1]{\href{https://orcid.org/#1}{\includegraphics[width=8pt]{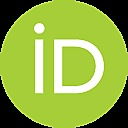}}}

\definecolor{backgG}{RGB}{255, 255, 153}
\definecolor{tagtxtG}{RGB}{102, 102, 0}
\definecolor{backgPc}{RGB}{179, 255, 179}
\definecolor{tagtxtPc}{RGB}{0, 102, 0}
\definecolor{backgPw}{RGB}{255, 179, 179}
\definecolor{backgPw}{rgb}{0.0, 1.0, 1.0}
\definecolor{tagtxtPw}{RGB}{0.0, 1.0, 1.0}
\definecolor{backgPo}{rgb}{0.0, 1.0, 1.0}
\definecolor{tagtxtPo}{RGB}{102, 0, 0}
\definecolor{backgPm}{rgb}{0.98, 0.81, 0.69}
\definecolor{tagtxtPm}{RGB}{0,1,1}

\begin{document}
%
\title{Can pre-trained language models generate titles for research papers?}

\author{Tohida Rehman\inst{1} \orcid{0000-0002-3578-1316} \and
Debarshi Kumar Sanyal\inst{2} \orcid{0000-0001-8723-5002} \and
Samiran Chattopadhyay\inst{1, 3} \orcid{0000-0002-8929-9605}}

\authorrunning{T. Rehman et al.}
\institute{Jadavpur University, Kolkata-700106, India.\\   \email{tohida.rehman@gmail.com}\\
\and
Indian Association for the Cultivation of Science, Kolkata-700032, India.\\
\email{debarshi.sanyal@iacs.res.in}\\
\and 
Techno India University, Kolkata-700091, India. 
\email{samirancju@gmail.com}
}

\maketitle              

\begin{abstract}
The title of a research paper communicates in a succinct style the main theme and, sometimes, the findings of the paper. Coming up with the right title is often an arduous task, and therefore, it would be beneficial to authors if title generation can be automated. In this paper, we fine-tune pre-trained language models to generate titles of papers from their abstracts. Additionally, we use GPT-3.5-turbo in a zero-shot setting to generate paper titles. The performance of the models is measured with ROUGE, METEOR, MoverScore, BERTScore and SciBERTScore metrics. We find that fine-tuned PEGASUS-large outperforms the other models, including fine-tuned LLaMA-3-8B and GPT-3.5-turbo, across most metrics. We also demonstrate that ChatGPT can generate  creative titles for papers. Our observations suggest that AI-generated paper titles are generally accurate and appropriate.

\end{abstract}
\section{Introduction}
\label{intro}
One of the most important aspects of writing a research paper is coming up with a concise and informative title. In academic publishing, a title should also be engaging, allowing readers to quickly grasp the paper's key insights. Even at first glance, readers should gain a clear understanding of the paper's contributions. An effective title includes keywords that describe the research, and sentences in the paper containing words from the title typically reflect its main theme. Many studies have been devoted to exploring the connection between the title of a paper and its readership as captured by citations and downloads \cite{jamali2011article,letchford2015advantage,rostami2014effect}. Short titles reportedly attract more citations \cite{letchford2015advantage}.

Given the tremendous success of neural language models, especially large language models (LLMs), in various natural language processing (NLP) tasks, it is natural to ask if titles of scientific papers can also be generated by these models. Given the expectation that a title should capture the key import of a paper and yet be a short sequence of words that appeal to a large readership, the task is a challenging one.  Title generation can be considered as a special case of abstractive text summarization, aiming to distill the most crucial information from a document.

In this paper, we aim to generate titles of research papers from their abstracts using pre-trained language models (PLMs). We fine-tuned several PLMs on a dataset of abstracts and titles. We use the term `LLM' to denote a PLM with more than a billion parameters. We experimented with both the pre-trained and fine-tuned versions of LLaMA-3 LLM. We have also used GPT-3.5-turbo in a zero-shot setting with prompts to generate titles from abstracts. We believe that machine generated titles could be very useful for non-native English speakers who find it difficult to quickly construct a suitable title for their papers, and also for novice researchers making their foray into the field. The generated titles can then be refined by them to better match their personal style and preference.
The main contributions of this paper are:
\begin{enumerate}
    \item We fine-tuned the PLMs \textbf{T5-base} \cite{JMLR:v21:20-074}, \textbf{BART-base} \cite{lewis-etal-2020-bart},  and \textbf{PEGASUS-large} \cite{10.5555/3524938.3525989} with \textbf{CSPubSum} dataset to automatically generate titles of research papers. We have also used the open-source LLM \textbf{LLaMA-3-8B} \cite{touvron2023llama, llama3modelcard} and contrasted its generation quality in two settings: with and without fine-tuning it for this specific task. Additionally, we have employed \textbf{GPT-3.5-turbo} in a zero-shot setting to generate titles from abstracts. 
    \item We have curated  a dataset of 1000 abstracts and titles, called \textbf{LREC-COLING-2024}, on which we evaluate our fine-tuned models without further training.
    \item We have measured the performance of the models using ROUGE \cite{lin2004rouge}, METEOR \cite{banerjee2005meteor}, MoverScore \cite{zhao-etal-2019-moverscore}, BERTScore \cite{zhang2019bertscore} and SciBERTScore metrics. We introduce SciBERTScore in this paper as a variant of BERTScore. We have also evaluated the factual accuracy of the model-generated titles at the entity level by employing metrics such as precision-source and F1-target, which were introduced in a prior study \cite{nan-etal-2021-entity}. Manual evaluation of a small subset of the generated titles has also been carried out.
    \item Our experiments show that \textbf{PEGASUS-large} outperforms the other models in terms of the chosen metrics. Notably, PEGASUS-large has far fewer parameters compared to LLMs like GPT-3.5-turbo and LLaMA.
    \item We have also analyzed whether ChatGPT can generate creative titles for research papers. In this analysis, we have compared the generated titles with the author-written titles and had them evaluated by human annotators.
    \item We have hosted a demo application on Render\footnote{\url{https://title-generation-researchpapers.onrender.com/}}, that allows a user to generate paper titles  using pre-trained language models. We have publicly released the fine-tuned models\footnote{\url{https://huggingface.co/TRnlp}} as well as the LREC-COLING-2024 corpus\footnote{\url{https://huggingface.co/datasets/TRnlp/LREC-COLING-2024-Abstract-Title}} on Hugging Face. Additional results are  available on our GitHub repository\footnote{\url{https://github.com/tohidarehman/Title-Generation-ResearchPapers}}.

\end{enumerate}

\section{Literature survey}
Automatic text summarization has a long history. 
As far back as 1958, Luhn et al. \cite{luhn1958automatic} pioneered an extractive summarization technique that selects sentences based on the frequency of significant words, excluding common words, to summarize technical papers and magazine articles. 
Lloret et al. \cite{lloret2013compendium} developed the COMPENDIUM system for generating biomedical research abstracts, utilizing both extractive and hybrid extractive-abstractive models to evaluate their effectiveness and suitability for summarization. 

The development of sequence-to-sequence models \cite{Sutskever-2014-sequence}, attention-based encoders with beam search and RNN \cite{bahdanau2015neural, nallapati2016abstractive}, and pointer-generator networks with coverage mechanisms \cite{See2017GetTT} have enriched the field of abstractive text summarization. These models have primarily been applied to news datasets. The next phase of research in summarization and natural language processing (NLP) in general was propelled by the transformer architecture \cite{vaswani2017attention}. It led to the development of various pre-trained language models 
such as T5 \cite{raffel2019exploring}, BART \cite{devlin2018bert}, and PEGASUS \cite{zhang2020pegasus}, that are initially trained in self-supervised manner on broad, general-purpose text datasets. Their extensive pre-training helps them capture linguistic patterns and knowledge. These pre-trained models can then be fine-tuned for specific tasks across various domains.

In the special case of title generation for a body of text, researchers have proposed methods to generate titles of news articles from their content. Tan et al. \cite{tan2017neural} proposed a coarse-to-fine approach, which first identifies the important sentences with hierarchical attention in a encoder-decoder framework. They tested their model using New York Times (NYT) and DUC-2004 dataset. Our perspective in this paper acknowledges that a title essentially acts as a summary of the content within the research paper \cite{nenkova2011automatic}. Mishra et al. \cite{mishra2021automatic} have proposed a method to generate titles for academic papers. First, they have generated a pool of candidate titles using a pre-trained model, then selected the most suitable one from them, and finally refined it to ensure semantic and syntactic accuracy, thereby enhancing the representativeness of the titles for the corresponding texts. They have used arXiv\footnote{\url{https://tinyurl.com/y9pu6xyp}}, ACL \cite{wang2018paper}, and ICMLA \cite{vallejo2019dataset} dataset. More recently, Liu et al. \cite{liu2022oag} pre-trained a transformer encoder, OAG-BERT, on computer science research  papers (including its metadata) and used it to produce paper titles, which were deemed acceptable by researchers. However, comprehensive evaluation of these generated titles using automatic metrics has not been done. Another emerging area of interest is the generation of research highlights from paper abstracts \cite{rehman2021automatic, rehman-etal-2022-named, rehman2023research, 10172215}. Distinct from the above, we use a diverse set of encoder-decoder and decoder-only transformers in this paper to generate titles from abstracts. We also perform a detailed evaluation of the generated titles. Our work aligns with the contemporary research that explores the potential of GPT and other artificial intelligence tools in education, research, and scholarly publishing \cite{lin2023and,ciaccio2023use}. 

\section{Datasets}
We have used the CSPubSum dataset provided by Collins et al.  \cite{collins2017supervised}, which contains URLs of 10147 computer science publications from ScienceDirect\footnote{\url{https://www.sciencedirect.com/}}. We crawled the dataset and organized each example as a pair of an \textit{abstract} and its corresponding \textit{title}. 
The dataset is split into 8120 examples for training, 1014 examples for validation, and 1013 examples for testing.
We crawled an additional set of 1000 examples from accepted papers at LREC-COLING 2024, that are available on ACL Anthology\footnote{\url{https://aclanthology.org/events/coling-2024/}} under CC BY 4.0 licence. This dataset, which we hereafter refer to as LREC-COLING-2024, consists of a pair of abstract and title for each paper. 

We train the selected language models exclusively on the train subset of CSPubSum but evaluate them both on the test subset of CSPubSum and the corpus of LREC-COLING-2024. While CSPubSum focuses on diverse topics of computer science, they primarily belong to the pre-transformer \cite{vaswani2017attention} era as the impact of transformers began to emerge after 2017, the year CSPubSum was released. We wanted to verify how well the trained models perform on a different and more recent dataset that they were not explicitly trained on. This led us to curate LREC-COLING-2024. The average number of tokens in a title in CSPubSum is 12 while it is 11 for LREC-COLING-2024, although a few titles are longer and some as small as two tokens. 
A few titles have multiple sentences; for example, one title reads: ``Comparative statics effects independent of the utility function. When do we act the same way under risk?''. 
However, around 82\% papers in CSPubSum and  90\% papers in LREC-COLING-2024 have no more than 15 tokens in their titles. Therefore, when fine-tuning or performing inference with the pre-trained models for title generation, we set the maximum title length to 20 tokens. This limit is also informed by the fact that shorter titles are generally more appealing to readers \cite{letchford2015advantage}.

\section{Methodology}
In this section, we detail the fine-tuning process of various PLMs. We have chosen the following models: 
\begin{enumerate}
    \item \textbf{T5-base} \cite{JMLR:v21:20-074}: It is an encoder-decoder model which is a slight variation of the original transformer model \cite{vaswani2017attention}. Formulating every text processing problem, including translation, question answering, and classification, as a ``text-to-text'' transformation problem, the same model is trained to perform these diverse tasks. To pre-train the model, random text spans are corrupted/dropped and the model is trained to generate them. T5-base contains 220M parameters. 
    \item \textbf{BART-base} \cite{lewis-etal-2020-bart}: A denoising autoencoder, it combines bidirectional and auto-regressive transformers exemplified by BERT \cite{devlin2018bert} and GPT \cite{brown2020language}, respectively. To train BART, the input text is first corrupted with a noising function, and then the model reconstructs the original text. It is particularly suitable for text generation problems. {BART-base} is configured with 139M parameters. 
    \item \textbf{PEGASUS-large} \cite{10.5555/3524938.3525989}: It is a transformer-based encoder-decoder model that is trained on large text collections with an objective function specifically focused on summarization. In particular, pre-training the model involves masking \textit{important} sentences from an input document and generating them as one output sequence. PEGASUS-large contains 568M parameters. 
    \item \textbf{LLaMA-3-8B}: We employed the pre-trained {LLaMA-3-8B} model\footnote{\url{https://ai.meta.com/blog/meta-llama-3/}}, which consists of 8 billion parameters. We used it in two ways: with fine-tuning and without fine-tuning. The LLaMA series of models \cite{touvron2023llama} are decoder-only Transformer-based large language models trained exclusively on publicly available datasets (in contrast to the GPT series). 

    \item \textbf{GPT-3.5-turbo}: We have used \texttt{gpt-3.5-turbo-1106}\footnote{\url{https://platform.openai.com/docs/models/gpt-3-5-turbo}}. Its web version is commonly known as ChatGPT-3.5\footnote{\url{https://chatgpt.com/}}. It is a decoder-only model pre-trained on massive text corpora and further fine-tuned using instruction tuning and reinforcement learning from human feedback. It is a successor to GPT-3, a large language model with 175B parameters. We used a prompt-based zero-shot in-context learning setup, where we simply prompted the model to generate a title given the abstract.
\end{enumerate}
We have fine-tuned the first four models on the train subset of CSPubSum. Note that although the above models come in multiple sizes, we have chosen the smallest versions that were freely available at the time of this research,  due to constraints on our computational resources. The paper that introduced the PEGASUS model \cite{10.5555/3524938.3525989} discusses both a base and a large model, but we could only find the checkpoints for {PEGASUS-large} on Hugging Face  and thus, we selected this version. Although GPT-4 is more powerful than GPT-3.5-turbo, the former is only available to paid subscribers, and therefore, we did not use it for this research. However, we anticipate that, as with other deep learning applications, using larger models will improve the quality of the generated titles.

\subsection{Data processing}
We removed extra spaces from the documents in both datasets and retained only the examples where the abstract length is at least 20 tokens and the paper title length is at least 3 tokens. Since we imposed a limit of 20 tokens on the generated title length, we stipulated the abstract to be longer than 20 tokens so that the task can be treated as a text summarization problem.

\subsection{Implementation details}
\label{sec:Impl}
We have chosen the following pre-trained models from the Hugging Face repository for fine-tuning on the CSPubSum dataset: {T5-base}\footnote{\url{https://huggingface.co/t5-base}}, {BART-base}\footnote{\url{https://huggingface.co/facebook/bart-base}}, {PEGASUS-large}\footnote{\url{https://huggingface.co/google/pegasus-large}}. 
Fine-tuning was performed for 5 epochs. We used  batch size of 32,  learning rate of 4e-5,  and  the {\tt metric\_for\_best\_model} as {\tt ROUGE1-F1}.
We used the following prompt to generate titles with GPT-3.5-turbo and LLaMA-3-8B LLMs:\\\\
\texttt{Create a concise title from this abstract using at most 20 tokens, \\highlighting the main contributions and focus. <ABSTRACT>}\\\\
During evaluation, both GPT-3.5-turbo and LLaMA-3-8B models were given the above prompt without any in-context examples. Unlike {GPT-3.5-turbo}, {LLaMA-3-8B} can be trained. But fine-tuning {LLaMA-3-8B} is an extremely computationally intensive task. Therefore, we used the parameter-efficient fine-tuning technique called Low-Rank Adaptation (LoRA). In particular, we have taken the {LLaMA-3-8B} \footnote{\url{https://huggingface.co/unsloth/llama-3-8b-bnb-4bit}} model from Hugging Face, then loaded it in 4-bit precision to save memory, and finally, fine-tuned it for 5 epochs using a learning rate of 4e-5, train batch size of 32 and eval batch size of 1, rank of the adaptation matrices {\tt r=16} , {\tt lora\_alpha = 16} (Scaling Factor), and using  {\tt peft\_config} for loading LoRA. For both fine-tuning and subsequent evaluation,  we utilized the same prompt mentioned earlier. For comparison, we also present results obtained with LLaMA-3-8B that is \textit{not} fine-tuned on this dataset; we denote this variant as {LLaMA-3-8B$^\ast$}. 
In case of {GPT-3.5-turbo}, we used a temperature setting of 0.3, to generate titles based on the abstracts.

For all models, we have set the maximum number of input tokens (i.e., the abstract length) to 512 and output tokens (i.e., the title length) to 20. In all cases, we have measured the memory and compute power consumption using the WandB tool\footnote{\url{https://wandb.ai/site}}. The fine-tuning of the {LLaMA-3-8B} model took 1 hour 43 minutes for 5 epochs. In the context of smaller PLMs, for 5 epochs of fine-tuning, {PEGASUS-large} model took 23 minutes, {BART-base} model took 7 minutes, and {T5-base} took 13 minutes. Training and fine-tuning of the models were done on Tesla A100-SXM4-40GB \texttt{Colab Pro+} that supports GPU-based training.

\subsection{Evaluation metrics}
We used the commonly used automatic text summarization evaluation metrics, including ROUGE \cite{lin2004rouge}, METEOR \cite{banerjee2005meteor}, MoverScore \cite{zhao-etal-2019-moverscore}, BERTScore \cite{zhang2019bertscore} and SciBERTScore \cite{beltagy2019scibert}, to assess the quality of the generated titles with reference to the author-crafted  titles.  ROUGE scores measure $n$-gram overlap between the generated titles and the ground-truth titles. We have used ROUGE-1, ROUGE-2, and ROUGE-L where the first uses unigram overlap, the second bigram overlap, and the last compares the longest common subsequence between the generated title and the golden title. METEOR measures sentence-level accuracy based on the alignment between the generated text and the reference text. In contrast, MoverScore, BERTScore, and SciBERTScore aim to measure the \textit{semantic similarity} between the output and the true title, using latent representations (embeddings) of the texts. MoverScore combines  Word Mover's Distance \cite{kusner2015word} and contextualized embeddings of the output and the ground-truth titles for semantic matching \cite{zhao-etal-2019-moverscore}. BERTScore calculates the cosine similarity between the BERT embeddings of the words in the two text sequences \cite{zhang2019bertscore}. Since the current application is in the domain of computer science, we have developed a variant of BERTScore, where we used SciBERT \cite{beltagy2019scibert} to generate the embeddings, then computed their cosine similarity as in BERTScore; we call the modified metric, SciBERTScore.

However, these metrics are inadequate to quantify \textit{factual consistency} \cite{kryscinski-etal-2020-evaluating}, which, in our context, refers to whether the entities present in the generated titles also occur in the corresponding abstracts and the extent to which these entities overlap with those in the author-written titles. Therefore, we have utilized three new metrics, introduced by \cite{nan-etal-2021-entity}, to assess the factual consistency of the generated titles. 

Let us call the author-written title as the target $t$, the model-generated title as the hypothesis $h$ and the input abstract as the source $s$. 
We define $\mathcal{N}(t)$ as the count of named-entities in the target (author-written title) and $\mathcal{N}(h)$ as the count of named-entities in the hypothesis (model-generated title). Then, $\mathcal{N}(h \cap s)$ is the number of entity matches between the generated title and the input source (abstract). Some named-entities in the title span multiple words. If even some words of a named entity in the title occur in the abstract, we consider it as a successful entity match. We used the scispaCy model \texttt{en\_core\_sci\_sm}\footnote{\url{https://allenai.github.io/scispacy/}} to identify entities.
\textbf{Precision-source}, defined as $prec_s= \mathcal{N}(h \cap s) / \mathcal{N}(h) $, 
is a metric used to assess the intensity of hallucination in relation to the \textit{input source (abstract)}. Note that 
$prec_s$ represents the percentage of entities mentioned in the generated title that can be retrieved from the input source document (abstract).  Low $prec_s$ indicates the possibility of  hallucination in the generated title. However, $prec_s$ does not capture the generated title's entity-level correctness in relation to the \textit{author-written title}. 
Therefore, recollecting that the author-written title is the target $t$, we define target-level entity accuracy in terms of \textbf{precision-target}  
$prec_t= \mathcal{N}(h \cap t) / \mathcal{N}(h);$    
 \textbf{recall-target}  
$recall_t = \mathcal{N}(h \cap t) / \mathcal{N}(t);$ and \textbf{F1-target}   
$F1_{t} = \frac{2*(recall_t*prec_t)}{recall_t+prec_t}.
$ 
Here, $\mathcal{N}(h \cap t)$ represents the number of matched named-entities in the generated title and the author-written title. 
Note that we have calculated the above mentioned \textit{precision-source} and \textit{precision-target} in two ways.
The first considers entity mentions in each document (which may be the source $s$ or target $t$ or hypothesis $h$) as a set so that multiple occurrences of an entity in a document are equivalent to a single occurrence. The other treats  the entity mentions in the input as a list. In this case, if a metric is defined as $\mu = \mathcal{N}(x \cap y) / \mathcal{N}(x)$, then for each entity mention in $x$, we check if it occurs in $y$; if it does, we increment the intersection count  $\mathcal{N}(x \cap y)$ by unity; this approach is followed in \cite{nan-etal-2021-entity}. In the first approach, we denote the metrics as $prec_s^{U}$, $prec_t^{U}$, $recall_t^{U}$, and $F1_t^{U}$ ($U$ indicates that only unique entity mentions are considered). In the second, we represent them as $prec_s^{NU}$, $prec_t^{NU}$, $recall_t^{NU}$, and $F1_t^{NU}$ ($NU$ denoting \textit{not unique}).

\section{Results}
\subsection{Quantitative comparison of different fine-tuned models}
\label{sec:quanResults}
In this sub-section, we report the results of experiments on the test dataset of {CSPubSum} dataset as well as {LREC-COLING-2024} dataset. Table \ref{Table:par_all_types_rouge_meteor_bert-cs} shows the performance in terms of ROUGE, METEOR, and semantic metrics like MoverScore, BERTScore and SciBERTScore, for the {CSPubSum} test dataset. Evaluation with entity-level factual consistency metrics precision-source($prec_s^{NU}$, $prec_s^{U}$),  precision-target ($prec_t^{NU}$, $prec_t^{U}$), recall-target ($recall_t^{NU}, recall_t^{U}$), and F1-target ($F1_t^{NU}$, $F1_t^{U}$) are shown in Table \ref{Table:entity-scoreCSPuBSum}. 

The overall finding is that \textbf{PEGASUS-large} fine-tuned on the CSPubSum dataset achieves the highest scores for all the above metrics except the precision-target metric.
Table \ref{Table:par_all_types_rouge_meteor_bert-coling} and Table \ref{Table:entity-scorecoling} show that the same {PEGASUS-large} model, that has been fine-tuned on the CSPubSum dataset, achieves the best performance on the {LREC-COLING-2024} dataset across all metrics except BERTScore. Thus, the superlative performance of {PEGASUS-large} in CSPubSum carries over to this dataset although it is not fine-tuned on it. 

We believe the reason {PEGASUS-large} wins the competition is that it is trained with an objective function that favors summarization, i.e., generating important lines that have been masked in the input document. Our observations show that although the LLMs show excellent performance on a wide range of NLP tasks, smaller models fine-tuned on custom datasets may exhibit better performance for a given task.

\begin{table*}[!h]
\centering
\caption{\small Evaluation of the quality of the generated titles in \textbf{CSPubSum}.}
\label{Table:par_all_types_rouge_meteor_bert-cs}
\begin{adjustbox}{width=1.0\linewidth}
{\begin{tabular}{|lccccccc|} \hline
Model Name &ROUGE-1 &ROUGE-2 &ROUGE-L &METEOR  &MoverScore &BERTScore &SciBERTScore\\\hline
T5-base &44.25 &25.04 &38.92 &38.36 &38.09 &89.9 &76.06\\ \hline
BART-base &45.7 &25.97 &40.11 &39.37 &39.75  &90.21 &76.89\\ \hline
PEGASUS-large &\bf{46.75} &\bf{27.13} &\bf{40.67} &\bf{42.61} &\bf{40.43}  &\bf{90.35} &\bf{76.93}\\ \hline
LLaMA-3-8B$^\ast$ &28.4 &12.58 &24.6 &27.17 &21.42  &86.34 &66.65\\ \hline
LLaMA-3-8B &40.8 &21.23 &36.57 &34.5 &37.02  &89.99 &76.41\\ \hline
GPT-3.5-turbo &42.81 &21.16 &36.55 &35.12 &37.39 &88.66 &76.28 \\ \hline
\end{tabular} }
\end{adjustbox}
\end{table*}

\begin{table*}[!h]
\centering
\caption{\small Evaluation of factual consistency of the generated titles in \textbf{CSPubSum}.}
\label{Table:entity-scoreCSPuBSum}
\begin{adjustbox}{width=0.98\linewidth}
{\tiny
{\begin{tabular}{|ccccccccc|} \hline

Model Name &$prec_s^{NU}$ & $prec_s^{U}$& $prec_t^{NU}$& $recall_t^{NU}$ & $F1_t^{NU}$ & $prec_t^{U}$& $recall_t^{U}$& $F1_t^{U}$ \\ \hline
T5-base &97.1 &97.08 &59.3 &51.82 &52.38 &59.08&51.58 &52.17\\ \hline
BART-base &97.44 &97.39 &\bf{61.52} &52.84 &53.91 &\bf{61.35} &52.56 &53.72  \\ \hline
PEGASUS-large &\bf{98.13} &\bf{98.08} &60.39 &\bf{56.49} &\bf{55.39} &60.17 &\bf{56.21} &\bf{55.18} \\ \hline
LLaMA-3-8B$^\ast$ &67.7 &67.7 &39.19 &40.48 &36.81 &38.96 &39.98 &36.48 \\ \hline
LLaMA-3-8B &78.95 &78.93 &57.54 &46.99 &49.12 &57.44&46.97 &49.08 \\ \hline
GPT-3.5-turbo &92.73 &92.73 &59.79 &50.95 &51.7 &59.79 &51.01 &51.74 \\ \hline
\end{tabular} 
}
}
\end{adjustbox}
\end{table*}

\begin{table*}[!h]
\centering
\caption{\small Evaluation of the quality of the generated titles in \textbf{LREC-COLING-2024}.}
\label{Table:par_all_types_rouge_meteor_bert-coling}
\begin{adjustbox}{width=1.0\linewidth}
{\begin{tabular}{|cccccccc|} \hline
Model Name &ROUGE-1 &ROUGE-2 &ROUGE-L &METEOR   &MoverScore &BERTScore &SciBERTScore \\\hline
T5-base &46.84 &28.7 &41.69 &39.88 &39.6  &88.71 &76.05\\ \hline
BART-base &46.87 &27.66 &41.89 &38.93 &39.61  &88.84 &75.94\\ \hline
PEGASUS-large &\bf{49.85} &\bf{30.51} &\bf{43.93} &\bf{43.23} &\bf{41.66}  &89.1 &\bf{76.74}\\ \hline
LLaMA-3-8B$^\ast$ &32.92 &16.66 &27.66 &30.61 &25.68  &86.77 &67.42 \\ \hline
LLaMA-3-8B &45.3 &26.53 &40.51 &38.18 &38.93 &88.83 &76.14\\ \hline
GPT-3.5-turbo &45.16 &23.97 &38.88 &37.45 &38.85  &\bf{89.54} &75.64\\ \hline
\end{tabular} }
\end{adjustbox}
\end{table*}

\begin{table*}[!h]
\centering
\caption{\small Evaluation of factual consistency of the generated titles in \textbf{LREC-COLING-2024}.}
\label{Table:entity-scorecoling}
    \begin{adjustbox}{width=.98\linewidth}
{\tiny
{\begin{tabular}{|ccccccccc|} \hline

Model Name &$prec_s^{NU}$ & $prec_s^{U}$& $prec_t^{NU}$& $recall_t^{NU}$ & $F1_t^{NU}$ & $prec_t^{U}$& $recall_t^{U}$& $F1_t^{U}$ \\ \hline
T5-base &97.44 &97.39&63.06 &59.24 &57.17 &62.8 &58.88 &56.87 \\ \hline
BART-base &96.24 &96.23 &64.3 &59.21 &57.78 &64.11 &58.92 &57.56  \\ \hline
PEGASUS-large &\bf{97.87} &\bf{97.85} &\bf{66.32} &\bf{64.64} &\bf{61.47} &\bf{66.15} &\bf{64.30} &\bf{61.28} \\ \hline
LLaMA-3-8B$^\ast$ &74.6 &74.52 &45.29 &48.16 &43.48 &45.02 &47.64 &43.13 \\ \hline
LLaMA-3-8B &81.57 &81.55 &62.18 &55.99 &55.19 &62.13 &55.95 &55.16 \\ \hline
GPT-3.5-turbo &91.98 &91.98 &60.87 &57.66 &55.56 &60.84 &57.69 &55.56 \\ \hline
\end{tabular} 
}
}
\end{adjustbox}
\end{table*}

Now, let us take a more nuanced view of the performance tables, beginning with the results on $n$-gram and semantic similarity between the author-written titles and the generated titles, as shown in Table \ref{Table:par_all_types_rouge_meteor_bert-cs}. It is immediately clear that the performance of {LLaMA-3-8B$^\ast$} (without fine-tuning) is considerably worse than that of the other models which are fine-tuned on  domain-specific data. This suggests that fine-tuning remains useful for attaining good performance despite the extensive pre-training that the models undergo. In contrast to LLaMA, GPT-3.5-turbo performs better, even in a zero-shot setting, possibly because it is continuously used worldwide and as a result gets continually trained. A second observation is that although {LLaMA-3-8B} (fine-tuned) and GPT-3.5-turbo have smaller ROUGE and METEOR scores compared to the smaller pre-trained models, their performance is closer to that of the latter according to the semantic metrics like SciBERTScore. For example, SciBERTScore values of all models except {LLaMA-3-8B$^\ast$} in Table \ref{Table:par_all_types_rouge_meteor_bert-cs} lie between 76.06 and 76.93. This indicates that all models (except {LLaMA-3-8B$^\ast$}) produce outputs with similar semantic similarity to the original titles. It also highlights the highly abstractive nature of the LLMs' outputs, which exhibit low word overlap with the original titles yet remain semantically close to them. Similar observations can be made from Table \ref{Table:par_all_types_rouge_meteor_bert-coling}. 

Now let us analyze the entity overlap between the generated titles and the author-written titles and abstracts. Table \ref{Table:entity-scoreCSPuBSum} displays the figures for {CSPubSum}. First note that the precision-source as well as F1-target do not vary much whether or not we count multiple occurrences of an entity. Again, {LLaMA-3-8B$^\ast$} (without fine-tuning) achieves the lowest scores for these metrics. The smaller pre-trained models display similar performance to one another. Consider precision source $prec_s^{NU}$: its large value indicates that the entities in the generated summary are mostly present in the input source which is the abstract here. $prec_s^{NU} = 97.1$ for {T5-base}, $97.44$ for {BART-base}, and $98.13$ for {PEGASUS-large}. But $prec_s^{NU}$ is 92.73 for GPT-3.5-turbo and 78.95 for LLaMA-3-8B, showing that these models generate novel words more frequently. In contrast, the F1-target ($F1_t^{NU}$) lies in the 50's for all models, meaning that there is only moderate overlap between the entities in the ground-truth title and the generated title -- thus, a generated titles does not match the golden title quite well but, at least for smaller models, it generally does not include entities outside the given abstract. Note that $F1_t^{NU}$ for LLMs is lower than that for the smaller models, again pointing to the abstractive nature of their output. We observe a similar pattern in Table \ref{Table:entity-scorecoling} which reports the entity overlap statistics for the {LREC-COLING-2024} dataset. Clearly, the lower values of $prec_s^{NU|U}$ and $F1_t^{NU|U}$ for LLMs could indicate hallucination. But upon manually examining a few generated titles we did not notice any instances of hallucination. This hints at the presence of novel words that preserve the intended meaning. 

\subsection{Case studies}
\label{sec:caseStudies}
We show three representative examples of title generation, one from {CSPubSum} (test subset) and two from {LREC-COLING-2024}  to illustrate the behavior of the models. 
Fig. \ref{fig:sample_Abs_title-cspubsum3} shows the example from {CSPubSum}: the fine-tuned {T5-base}, {BART-base}, and {PEGASUS-large} models have generated titles that have significant similarity with the author-written titles. However, {T5-base} has generated a very long title. {BART-base} and {PEGASUS-large} generated the same titles. 
In contrast, {LLaMA-3-8B} {without fine-tuning}, produced a title that resembled a one-sentence summary and even that is incomplete; it is more of an extract from the abstract. After {fine-tuning}, the {LLaMA-3-8B} model generated an acceptable title. GPT-3.5-turbo, using prompt-based techniques, successfully generated a title that also captures the essence of the paper and is similar to other generated titles and the author-written title. 
Recollect that according to the evaluation metrics in Table \ref{Table:par_all_types_rouge_meteor_bert-cs}, the titles generated by {GPT-3.5-turbo} score lower than those generated by the fine-tuned {PEGASUS-large} model, despite the former's stylistic flair. This discrepancy highlights a limitation of the automated metrics in accurately evaluating titles generated in a highly abstractive and stylistic manner.
\begin{figure*}[!htb] 
 \centering 
 \begin{tabular}{|p{12.5 cm}|} \hline
 {\bf Author-written title:} \textcolor{blue}{``Comparative statics effects independent of the utility function. When do we act the same way under risk?''} 
\\\hline
 {\bf T5-base:} ``Comparative statics effects independent of the utility function for portfolio choice and competitive firm under price uncertainty''\\\hline 	    
 {\bf BART-base:} ``Comparative statics effects in the context of expected utility''\\\hline
 {\bf PEGASUS-large:} ``Comparative statics effects in the context of expected utility''\\\hline
 {\bf  LLaMA-3-8B$^*$:} ``The author proposes a methodological approach that enables comparative static analysis of various economic models, irrespective of the''\\\hline
 {\bf LLaMA-3-8B:} ``Comparative statics analysis: A new approach''\\\hline
 {\bf Zero-shot GPT-3.5-turbo:} ``Comparative statics effects on expected utility in decision-making''\\\hline
 \end{tabular} 	
 \caption{\small Input is an abstract from \textbf{CSPubSum}. Titles generated by the different models are shown.  Paper  taken from \texttt{\small \url{ https://www.sciencedirect.com/science/article/abs/pii/S037722171500586X}}.}	
 \label{fig:sample_Abs_title-cspubsum3} 
 \end{figure*}

Now let us look at Fig. \ref{fig:sample_Abs_title-coling-1}, which shows the outputs generated by the models given an abstract from the {LREC-COLING-2024} dataset. In case of {BART-base}, the generated title is incomplete. This is due to the hard limit on the output token count. We have observed this problem with other models, too. A simple workaround (without retraining the models) is to increase the output token limit during inference, which sometimes leads to a grammatically correct and semantically complete title. While {T5-base} and fine-tuned {LLaMA-3-8B} have generated acceptable titles, they missed the ``low-resource'' setting which is accurately captured by {PEGASUS-large}. {GPT-3.5-turbo} generated almost the same title as that of {PEGASUS-large}, except that it uses the abbreviation ``NMT'', and abbreviations may not be desirable in a title. 

\begin{figure*}[!htb] 
\centering 
\begin{tabular}{|p{12.5 cm}|} \hline
{\bf Author-written title:} \textcolor{blue}{``A Reinforcement Learning Approach to Improve Low-Resource Machine Translation Leveraging Domain Monolingual Data''} 
\\\hline
{\bf T5-base:} ``Reinforcement Learning Domain Adaptation for Neural Machine Translation''\\\hline 	    
{\bf BART-base:} ``A novel Reinforcement Learning Domain Adaptation method for Neural Machine Translation in the low-''\\\hline
{\bf PEGASUS-large:} ``Reinforcement learning domain adaptation for Neural Machine Translation in the low-resource domain''\\\hline
{\bf  LLaMA-3-8B$^*$:} ``Reinforced Domain Adaptation Method for Low Resource Neural Machine Translation''\\\hline
{\bf LLaMA-3-8B:} ``Reinforcement learning domain adaptation for neural machine translation''\\\hline
{\bf Zero-shot GPT-3.5-turbo:} ``Reinforcement Learning Domain Adaptation for Low-Resource NMT''\\\hline
\end{tabular} 	
\caption{\small Input is an abstract from \textbf{LREC-COLING-2024}. Titles generated by the different models are shown. Paper taken from   \texttt{\url{https://aclanthology.org/2024.lrec-main.132/}}.}	
\label{fig:sample_Abs_title-coling-1} 
\end{figure*}

In the example shown in Fig. \ref{fig:sample_Abs_title-coling-2}, {PEGASUS-large} and {fine-tuned LLaMA-3-8B} generated titles that are identical to the author-written title. A quick reading of the abstract of the input paper indicates that {T5-base} has captured an important aspect of the proposed model in the paper, namely the ``entity abstraction approach''; however, the model in the paper has other characteristics and singling out one may not be appropriate. The title generated by {BART-base} is logically incomplete as it misses the phrase ``for entailment tree generation''. {LLaMA-3-8B without fine-tuning} struggled to generate a succinct title. In contrast, GPT-3.5-turbo generated an interesting title which captures the very purpose of the proposed model: ``improving AI explanations''.

\begin{figure*}[!htb] 
 \centering 
 \begin{tabular}{|p{12.5 cm}|} \hline
 {\bf Author-written title:} \textcolor{blue}{``A Logical Pattern Memory Pre-trained Model for Entailment Tree Generation''} 
\\\hline
 {\bf T5-base:} ``An entity abstraction approach for logical pattern memory pre-trained models''\\\hline 	    
 {\bf BART-base:} ``Logical pattern memory pre-trained model''\\\hline
 {\bf PEGASUS-large:} ``A logical pattern memory pre-trained model for entailment tree generation''\\\hline
 {\bf  LLaMA-3-8B$^*$:} ``The proposed method addresses the limitations of previous approaches
by incorporating an external memory structure to capture the latent representations''\\\hline
 {\bf LLaMA-3-8B:} ``Logical pattern memory pre-trained model for entailment tree generation''\\\hline
 {\bf  Zero-shot GPT-3.5-turbo:} ``Improving AI Explanations with Logical Pattern Memory Pre-trained Model''\\\hline
 \end{tabular} 	
 \caption{\small Input is an abstract from \textbf{LREC-COLING-2024}. Titles generated by the different models are shown.  Paper  taken from \texttt{\small \url{ https://aclanthology.org/2024.lrec-main.68/}}.}	
 \label{fig:sample_Abs_title-coling-2} 
 \end{figure*}

\subsection{Manual evaluation}
\label{sec:manualEval}
We selected twenty papers, ten from each dataset (for CSPubSum, they were sampled from the test subset) for human evaluation. A human annotator working in the field of NLP was asked to choose the most appropriate title among the  ones generated by the six models. It was found that for CSPubSum, in $80\%$ of the cases, the titles generated by \textbf{PEGASUS-large} were the most preferred. 
For the LREC-COLING-2024 dataset, \textbf{PEGASUS-large} was the winning model in 50\% of cases, while the titles generated by GPT-3.5-turbo were considered the best in 40\% of cases.

\subsection{Creative titles with LLMs}
We explored whether  ChatGPT can generate creative titles for papers. In particular, we prompted ChatGPT-4o to generate titles in specific styles -- namely,  \texttt{catchy}, \texttt{rhyming}, \texttt{funny}, and \texttt{comprehensive} -- one style at a time for ten papers from CSPubSum (test set) and ten from LREC-COLING-2024. An example is presented in Fig.   \ref{fig:sample_Abs_title-coling-3}, which displays titles in the above four  styles along with the `plain' title generated by GPT-3.5-turbo (using the prompt in Section \ref{sec:Impl}) and the author-written one. We depict the similarity of these titles with the author-written ones in Table \ref{Table:creative_auto_eval} (ignore the last column). Catchy, comprehensive, and plain titles exhibit similar similarity scores. However, rhyming and  funny titles generally share very few words (sometimes with zero bi-gram overlap) and only moderate semantic similarity with the author-written titles. This is not surprising, as these styles often introduce novel words not present in the original titles, and tend to sacrifice technical terms to adhere to the stylistic constraints. 

\begin{figure*}[!htb] 
\centering 
\begin{tabular}{|p{12.5 cm}|} \hline
{\bf Author-written title:} \textcolor{blue}{``A Tool for Determining Distances and Overlaps between Multimodal Annotations''} 
\\\hline
{\bf Catchy:} ``Analyzing ELAN Annotations: A Tool for Comparing Multimodal Overlaps in Corpus Analysis''\\\hline 	    
{\bf Rhyming:} ``Comparative Insights in ELAN's Light: A Tool for Annotations That Delivers Clear Sight''\\\hline
{\bf Funny:} ``Annotation Adventures: Finding Overlaps and Other Mischief!''\\\hline
{\bf Comprehensive:} ``Tool for Analyzing ELAN Annotations: Comparing Speech and Gesture Overlap in Multimodal Corpora''\\\hline
{\bf Plain:} ``Tool for Comparing ELAN Annotations in Multimodal Corpus Analysis''\\\hline
\end{tabular} 	
\caption{\small Titles generated by ChatGPT for an abstract in \textbf{LREC-COLING-2024}. Paper taken from  \texttt{\url{https://aclanthology.org/2024.lrec-main.150/}}.}	
\label{fig:sample_Abs_title-coling-3} 
\end{figure*}

\begin{table*}[!h]
\centering
\caption{\small Evaluation of ChatGPT-generated creative titles.} \label{Table:creative_auto_eval}
    \begin{adjustbox}{width=1.0\linewidth}
{\begin{tabular}{|c|cccccccc|c|} \hline
Dataset~ & Style &ROUGE-1 &ROUGE-2 &ROUGE-L &METEOR   &MoverScore &BERTScore &SciBERTScore & Avg. Rating\\\hline
\textbf{CSPubSum}~ & Catchy&45.81 &18.44 &38.83 &45.55 &35.43 &88.00  &75.77 & 4\\ \cline{2-10}
         & Rhyming&13.20 &0 &11.37 &10.37 &12.32 &83.47 &61.02 & 3.3\\ \cline{2-10}
         & Funny&18.57 &28.18 &14.95 &15.09 &14.76 &83.78 &59.53 & 2.8\\ \cline{2-10}
         & Compreh.& \textbf{48.03} & \textbf{29.32} & 40.88 &\textbf{47.93} &\textbf{44.21} &\textbf{88.16} &76.78 & \textbf{4.4}\\ \cline{2-10}
         & Plain&47.51 &25.03 &\textbf{42.09} &44.99 &40.74 &88.15 &\textbf{77} & 3.4\\ \hline\hline
\textbf{LREC-COLING-2024}~ & Catchy&47.82 &\textbf{24.47} &\textbf{41.60} &45.61 &42.84 &90.07  &75.77 & 3.8\\ \cline{2-10}
                 & Rhyming &20.48 &02.65 &16.94 &16.64 &20.72 &84.37  &61.02 & 3.1\\ \cline{2-10}
                 & Funny&15.71 &0.0 &11.36 &10.11 &11.47 &84.05  &59.53 & 3.6\\ \cline{2-10}
                 & Compreh.&\textbf{48.22} &21.88 &41.12 &\textbf{48.83} &\textbf{45.71} &90.4  &76.78 & \textbf{4.1}\\ \cline{2-10}
                 & Plain&43.83 &19.45 &39.76 &39.42 &42.48 &\textbf{90.88}  &\textbf{77.00} & 3.7\\ \hline
\end{tabular} }
\end{adjustbox}
\end{table*}

We selected nine human annotators, all undergraduate or graduate students in  computer science, to rate the ChatGPT-generated titles on a scale of 1-5. The average ratings are reported in the last column of  Table \ref{Table:creative_auto_eval}. Comprehensive titles have scored the highest. Author-written titles for both datasets received an average rating of 4. When asked to choose the best title for each paper, the annotators preferred ChatGPT-generated titles to the author-written ones for nine LREC-COLING-2024 papers and seven CSPubSum papers, mostly favoring the comprehensive or catchy titles.

\section{Demo}
\begin{figure*} [!htbp] 
\centering
\includegraphics[width=0.98\linewidth]{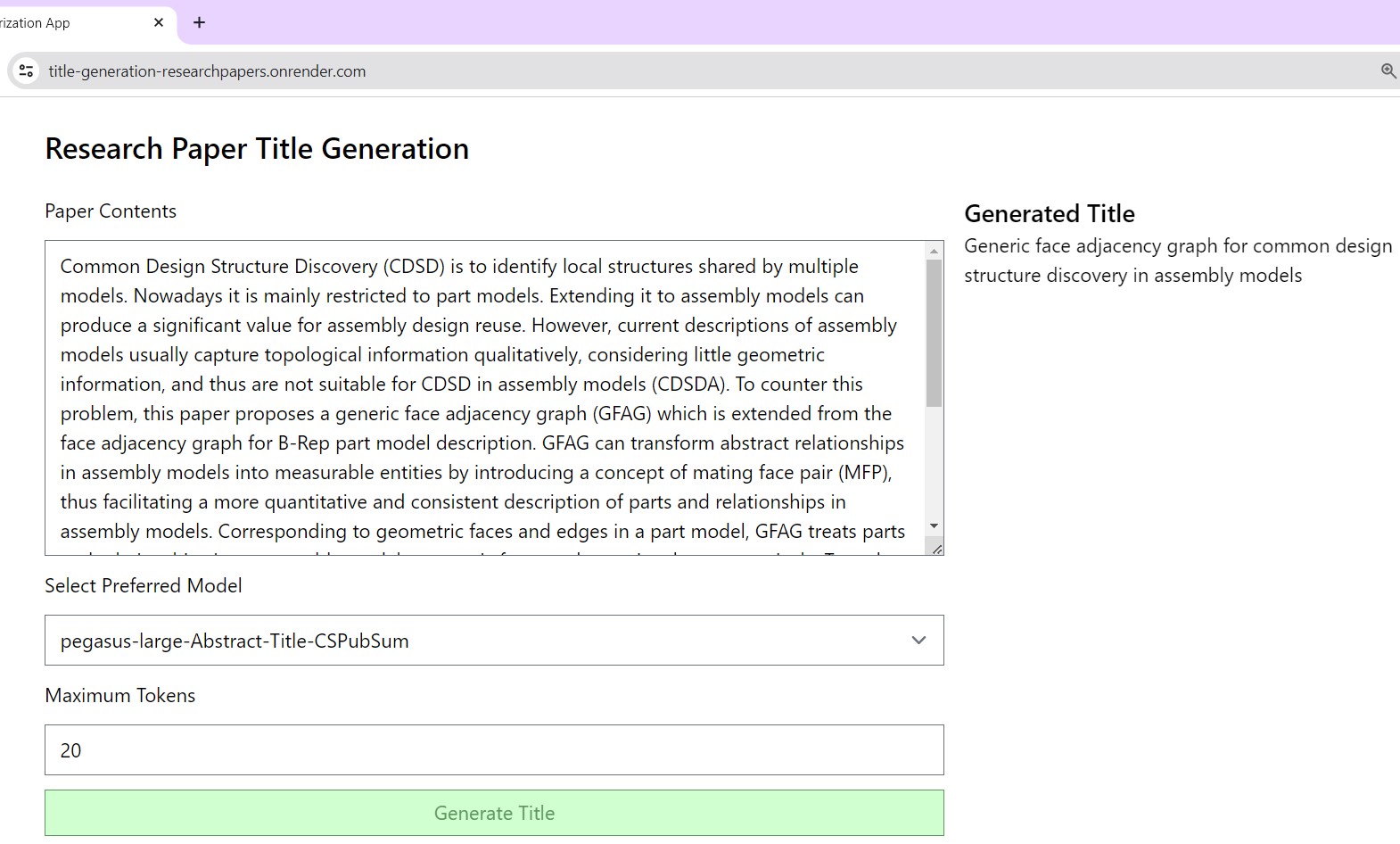}
\caption{\small Graphical user interface of our title generation application.}
\label{fig:model_diagram}
\end{figure*}

A demo of the application is publicly hosted. As shown in Fig.   \ref{fig:model_diagram}, a user can input an abstract into a text box, select a suitable fine-tuned language model and the maximum token count for the title, and obtain a title generated by the model. We have found that if the generated title is incomplete, increasing the token count (to say, 25 or 30) generally produces a correct and complete title.

\section{Conclusion}
We applied six different pre-trained language models to generate titles of research papers. The fine-tuned PEGASUS-large model has achieved the best performance on the CSPubSum dataset as well the LREC-COLING-2024 dataset. Our extensive experiments indicate that these language models can generate engaging and suitable titles for research papers. The success of PEGASUS-large suggests that lighter pre-trained language models, fine-tuned on domain-specific datasets, may be satisfactory and even more suitable for this task than large language models with billions of parameters and substantial computational requirements. The high scores achieved by the models on LREC-COLING-2024 dataset, on which they were not fine-tuned, indicates that fine-tuning on a related scholarly dataset suffices to generate acceptable performance. Although the generated titles are generally of high quality, authors might refine them further to better suit their papers. This tool can be particularly useful for non-native English speakers and novice researchers. In future, we would like to conduct a more detailed analysis of the difference between author-written and AI-generated titles, as well as  explore  better techniques for evaluating the abstractive quality of the outputs from LLMs.
\bibliographystyle{unsrt}
\bibliography{anthology}

\end{document}